\documentclass[letterpaper, 10 pt, conference]{ieeeconf}  

\IEEEoverridecommandlockouts                              

\usepackage{blindtext}
\usepackage{graphicx}

\usepackage[font=small]{caption}
\usepackage{amssymb}
\usepackage[cmex10]{amsmath}
\usepackage{float}
\usepackage{subcaption}
\usepackage{ tipa }
\usepackage{hyperref}
\usepackage[table]{xcolor}
\usepackage{mathpazo} 
\usepackage[final]{pdfpages}
\usepackage{csquotes} 
\usepackage{cite} 
\usepackage{siunitx} 
\usepackage{scrextend} 
\usepackage{courier} 
\usepackage{placeins} 
\usepackage{pgffor} 
\usepackage{csquotes}
\usepackage{tabularx}
\usepackage[english]{babel} 
\usepackage{blindtext}
\usepackage{array} 
\usepackage{mdwmath} 
\usepackage{eqparbox} 
\renewcommand{\thetable}{\arabic{table}}

\let\labelindent\relax 
\usepackage{enumitem}

\usepackage{hyperref} 
\makeatletter
\let\NAT@parse\undefined
\makeatother

\newcommand{\vect}[1]{\boldsymbol{#1}} 

\begin{document}
	%
	\title{I Can See Your Aim: Estimating User Attention From Gaze For Handheld Robot Collaboration}
	%
	%
	%
	\author{Janis Stolzenwald and Walterio W. Mayol-Cuevas \thanks{ 
			Department of Computer Science, University of Bristol, UK,
			{\color{white}....} 
			janis.stolzenwald.2015@my.bristol.ac.uk, wmayol@cs.bris.ac.uk
		}
	}
	\maketitle
	\begin{abstract}
		This paper explores the estimation of user attention in the setting of a cooperative handheld robot --- a robot designed to behave as a handheld tool but that has levels of task knowledge. We use a tool-mounted gaze tracking system, which after modelling via a pilot study, we use as a proxy for estimating the attention of the user. This information is then used for cooperation with users in a task of selecting and engaging with objects on a dynamic screen. Via a video game setup, we test various degrees of robot autonomy from fully autonomous, where the robot knows what it has to do and acts, to no autonomy where the user is in full control of the task. Our results measure performance and subjective metrics and show how the attention model benefits the interaction and preference of users.    
	\end{abstract}

	\IEEEpeerreviewmaketitle

	\vspace{-1em}
	\section{Introduction}
	Handheld robots are a new category of cooperative robots which are endowed with the look and behaviour of a handheld tool, include extra mechanical competences and process task information for better augmentation.
	
	Early results in this area demonstrate how a robot's task knowledge can be used for augmentation even without explicit feedback to the user via motion gesturing \cite{GreggSmith:2015bh} and with combinations of visual feedback \cite{GreggSmith:2016hn}. While informing the user of the robot's intention significantly improved task performance, a remaining challenge was identified: the conflict between user intention and the robot's plans. 
	This led to frustration in users and a negative impact on cooperative work measures. The studies particularly yield that this problem is rooted in unidirectional intention communication i.e. the robot displays its aim without observing the user. 
	Furthermore, when the robot operates under idealised full autonomy, its planning and performance tend to have high efficiency and its actions dominate those of the user. This has conflicted with users' perception of cooperative task solving e.g. sometimes there was a confusion about what the robot will do next.
	Our motivation stems from aiming to address the above issues and here we start by looking at incorporating models of user attention so that we can both: i) provide the robot with user's information and a proxy for her/his intention and ii) allow us to evaluate instances of conflicts between user and robot plans.
	
	Remote gaze tracking is a natural choice for this due to the extensive body of work linking gaze and action prediction. Land et al. found how eye gaze is closely related to a person's focus of attention as it precedes the location of actions during every-day tasks \cite{Land:2016kw}.
	
	\begin{figure}[t]
		\centering
		\includegraphics[width=0.629\linewidth]{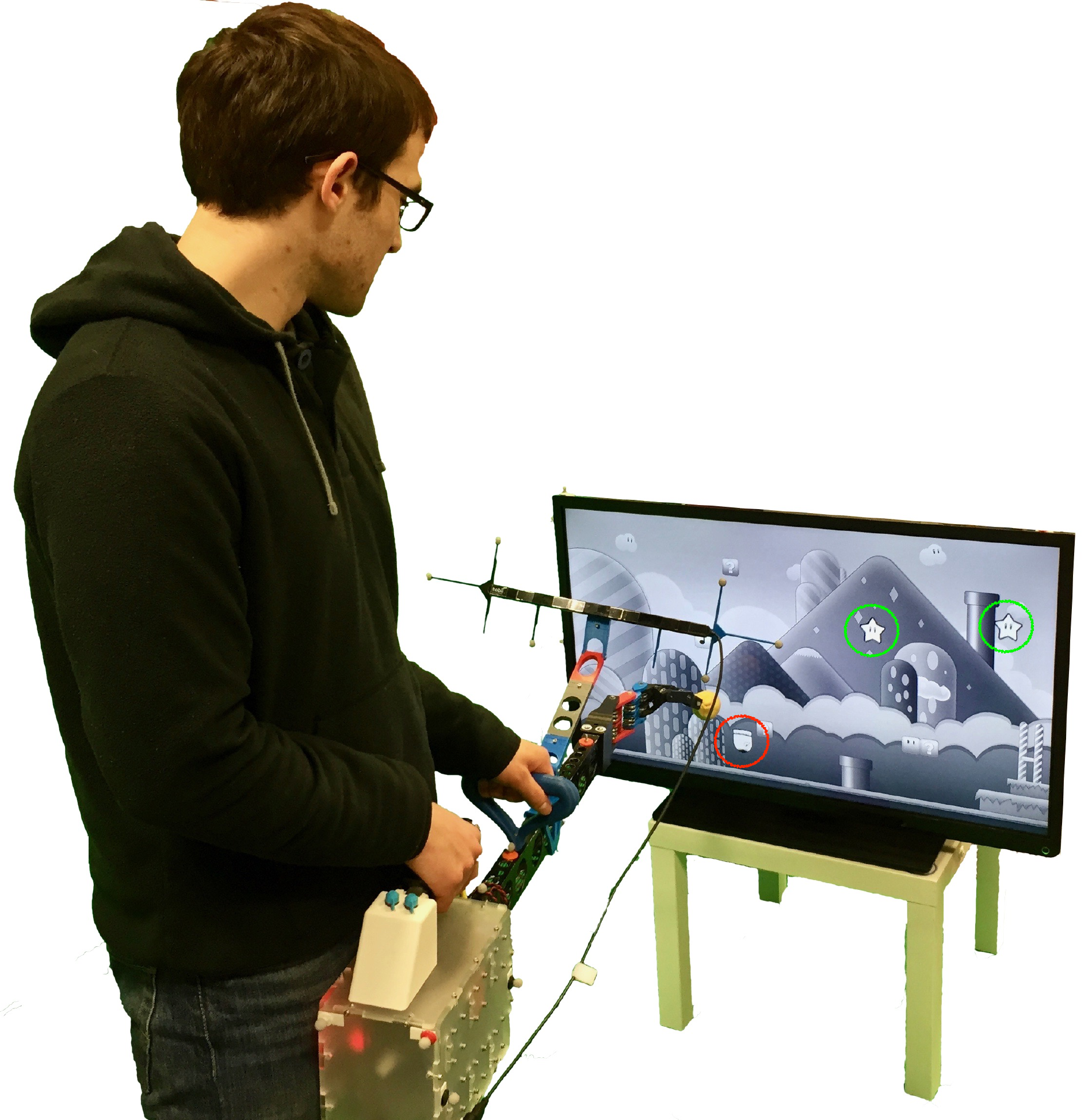}
		\caption{Testing setup for the interactive game. The user has to stop the targets (red) using the robot to increase the game score while distractors (green) can pass.}
		\label{fig:invgame}
	\end{figure}    

	For our study we use the open robotic platform platform\footnote{3D CAD models available from handheldrobotics.org} reported in \cite{GreggSmith:2016cz}, and modify it with the incorporation of a remote gaze tracker as described further down. The gaze information is first modeled and then used to evaluate its utility in a gamified cooperative task.
	This paper is formed by two principal parts, one that looks at the gaze tracker modeling and characterization and then the description and evaluation of the cooperative task under different modes of autonomy. 
	We conclude our paper with a discussion and summary.
	

	\section{Background and Related Work}
	Within this section, we review recent work about handheld robots and related fields such as wearables and intelligent tools. We also discuss means of attention estimation from past work which inspired our solution for the handheld robot. 
	
    \begin{figure}[t]
		\centering
		\includegraphics[width=0.99\linewidth]{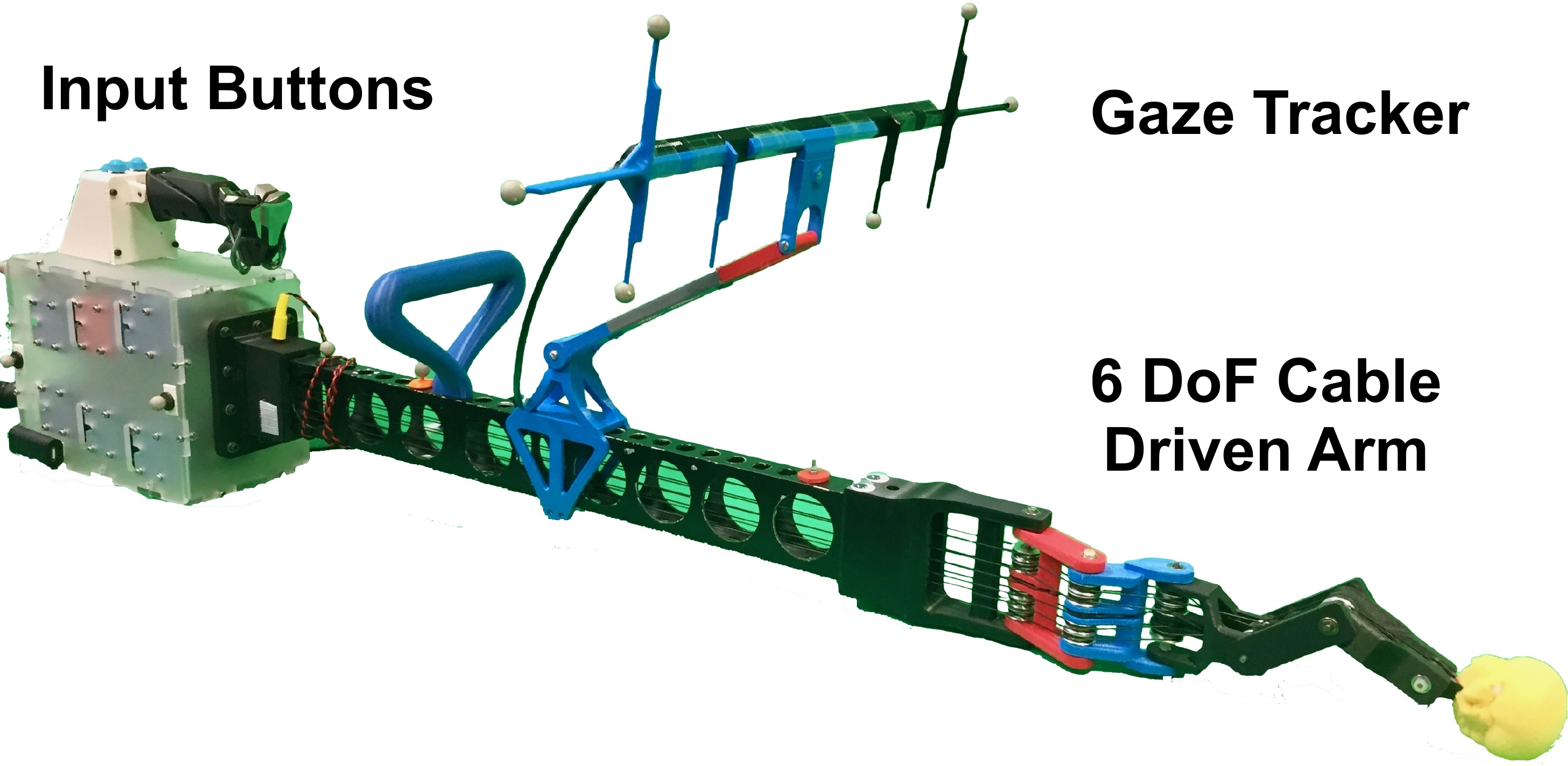}
		\caption{Handheld robot with extended user perception capabilities through a newly integrated eye tracking system.}
		\label{fig:front-profile-robot-labeled}
		\vspace{-0.5em}
	\end{figure}
	We pose the following research questions:
	\begin{enumerate}[label=\textbf{Q\arabic*}]
		\item How can user attention be used to enhance cooperation with handheld robots?\label{q1}
		\item How does the incorporation of attention affect task performance and the user's perceived task load?\label{q2}
	\end{enumerate}
		
	\subsection{Handheld Robots}
	The notion of non-medical, generic handheld robotics was proposed recently in \cite{GreggSmith:2015bh}. In that instance, a lightweight device with a cable driven trunk-shaped actuated arm that can move with 4 DoF while being tactically moved by a human was introduced. The tool itself is aware of the task and its progress. Therefore, it can augment the user during task execution. Furthermore, it can provide task related guidance using its 4-DoF end effector to point towards the goal. The robot was evaluated through experimental feasibility studies. The results show that an increase in the level of autonomy improves critical aspects of efficiency such as time-to-complete and perceived workload. At the same time, increased autonomy led to frustration in some participants which was expressed via statements like: \textit{The tool won't go where I want it to}. 
	
	In subsequent work, the lack of user guidance was addressed and different types of visual feedback e.g. through rudimentary robot gesturing, 2D-displays and stereoscopic VR-headsets were explored \cite{GreggSmith:2016cz,GreggSmith:2016hn}. The effect of the different feedback methods on the task performance was investigated through a comparative user study where participants were asked to complete a 5 DoF reaching task. The results yield that visual feedback improves task performance. Furthermore, it was found that participants could perform the task more accurately and perceived less workload when the robot was used compared to a manual completion of the same job.
	
	
	\subsection{Related Work}
	A concept that is closely related to the one of the handheld robot is the elbow mounted robotic forearm introduced by Vatsal and Hoffman \cite{Vatsal:2017dy}. The system shares the cooperative character with handheld robots but is distinguished by a much higher physical proximity which makes it more spatially dependant on the user.
    The results indicate that such a device could be purposed for industrial applications such as construction. At the same time, the authors suggest that a fluent cooperation with the robot would require a partially autonomous behaviour for the robot with an integrated user intention model.    
	
	Another close relative to handheld robots is a group of intelligent handheld tools, mainly purposed for industrial applications which augment users through visual guidance or mechanically correct their actions. 
	The research by Echtler et al. is about an intelligent welding gun that guides its user through the task, using augmented reality graphics on an LCD screen  \cite{Echtler:2003uo}. 
	Another example of a task aware tool is introduced by Rivers et al. \cite{Rivers:2012ff}. Their milling tool for 2D fabrication provides an on-display trajectory guidance and corrects the micro scale of the path while the rough positioning is done manually by the user.

	We note that on the above work there is little to no evidence of aiming to understand intention or user attention to enhance the cooperation between robot and user.
	
	\subsection{Behaviour Cues for Attention Estimation }
	Answering the question of how to best estimate a user's focus of attention in handheld robot task set-ups is the central aim of our research project.     
	The literature of using eye and/or head gaze for attention estimation is vast and spans many years. Recently, standard methods to detect gaze directions commonly involve eye trackers \cite{Mansouryar:2016uv} or determining head orientation \cite{Leelasawassuk:2015cq,Stiefelhagen:2004gr}. The information about both head orientation and eye gaze has been linked to a person's focus of attention in the past \cite{Odobez:2007hk, Mulvey:2012di}. 
	
	Land et al. found that there is a link between object interaction and preceding eye gaze during manual task execution \cite{Land:2016kw,Land:2001hl}. This is supported by Hayhoe et al. who studied hand-eye coordination and found that there is a small share of saccades on objects which were used shortly after these \textit{look ahead fixations} \cite{Anonymous:2001fx}. This observation was later linked to \textit{just-in-time} task planning by Mannie et al. who suggest that visual attention is turned towards a subsequent goal right before interaction rather than relying on spatial memory only \cite{Mennie:2006fo}.
	
	
	
	\section{Eye Tracking for a Handheld Robot}
	
	The aim of the integration of an eye tracking system is to gain attention-relevant gaze information of the user while the handheld robot is operated. In this section, we describe how a 3D gaze ray is constructed by merging 2D in-plane gaze information detected by a remote eye gaze tracker with support from motion capturing. 
	
	We use a Tobii Eye Tracker\footnote{https://help.tobii.com/hc/en-us/articles/213414285-Specifications-for-the-Tobii-Eye-Tracker-4C} which delivers gaze information such as the user's current eye position and the 2D intersection between eye gaze and screen surface in screen centre coordinates.
	
\begin{figure}[t]
		
		\vspace{1em}
		\centering
		\includegraphics[width=0.9\linewidth]{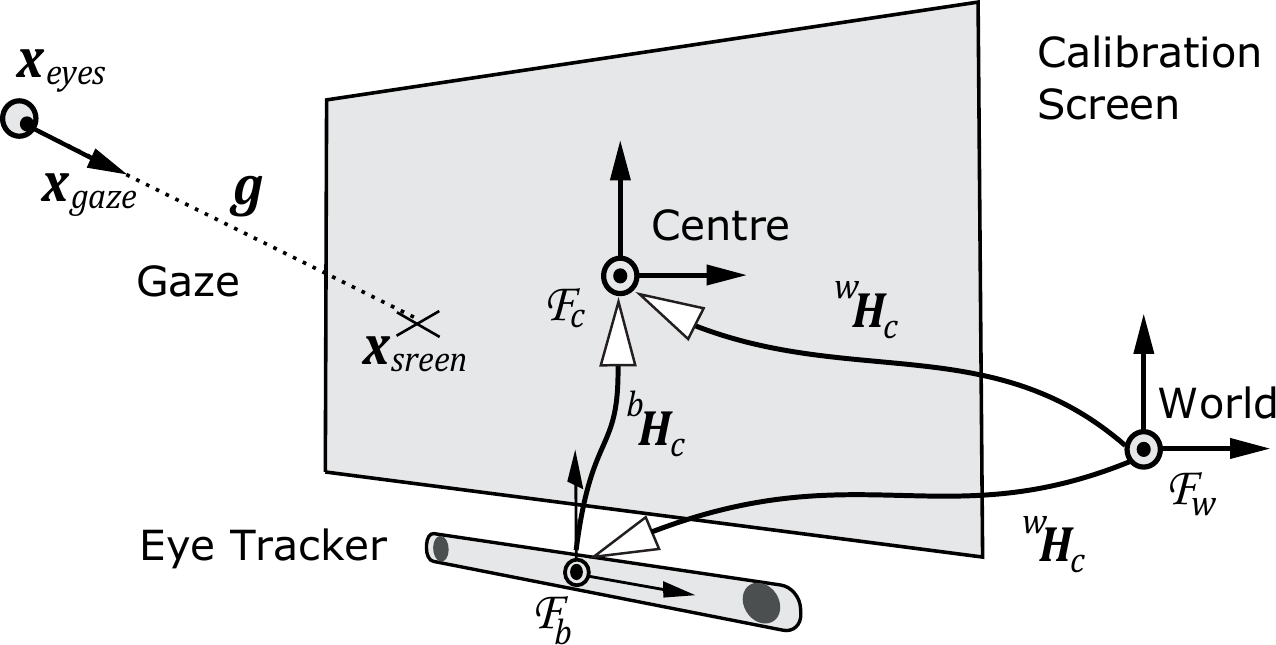}
		\caption{ This figure illustrates the calibration setup which is used to determine the (white tipped) transformation $^b\vect{H}_c$ which is required for the construction of the user's gaze $\vect{g}$.}
		\label{fig:as-frame-calibration}
	\end{figure}

	For the following, let $\mathcal{F}_w, \mathcal{F}_b, \mathcal{F}_c$ be the frames of the world, the tracker's base and the centre of a screen plane, respectively. Furthermore, let $^ix_{eyes}$, $^ix_{gaze}$ and $^ix_{screen}$ be vectors describing the eye's position, the gaze direction and the gaze intersection with a screen plane in the according frames ($i = w,b,c$). Consequently, $^i\vect{H}_j$ represents the homogeneous transformation between two frames $\mathcal{F}_i,\mathcal{F}_j$ \cite{Briot:2015jo}. The relationship between the frames can be seen in figure \ref{fig:as-frame-calibration}.
	
	Both the eye tracker's base and the calibration screen are coupled to the motion tracking system, hence $^w\vect{H}_b$ is known at any time, $^w\vect{H}_c$ is solely known during the offset calibration process while $^b\vect{H}_c$ is initially unknown. As the tracking device delivers the eye-positions and gaze intersection with an associated screen, we can obtain the local gaze $^c\vect{g}$ with respect to  $\mathcal{F}_c$ as follows\cite{Anonymous:PVm7qwCu}:
	
	\vspace{-1em}
	\begin{equation}
	^c\vect{g}(\lambda) = \,^c\vect{x}_{eyes} + \lambda ^c\vect{x}_{gaze}
	\end{equation}
	where $^c\vect{x}_{eyes}$ is the mean of the two eye positions and $^c\vect{x}_{gaze}$ is the gaze direction which can be derived by
	\begin{equation}
	\vect{x}_{gaze} = \frac{\vect{x}_{screen}-\vect{x}_{eyes}}{|\vect{x}_{screen}-\vect{x}_{eyes}|}
	\end{equation}
	
	Furthermore, while the eye tracker is attached to the calibration screen, we store the transformation $^b\vect{H}_c$ between tracker base and screen which is know through the equation:
	\vspace{-1em}
	\begin{equation}
	^b\vect{H}_c = (^w\vect{H}_c)^{-1} \,  ^w\vect{H}_b \,  
	\end{equation}
	Now, we want to use the eye tracker without it being attached to the screen which means we lose information about $^w\vect{H}_c$.
	However, it can be derived from combining the tracker's base with the stored transformation
	\begin{equation}
	^w\vect{H}_c = \,  ^w\vect{H}_b \,  ^b\vect{H}_c
	\end{equation}
	and finally, the gaze in world coordinates can be calculated in real time
	\vspace{-1em}    
	\begin{equation} \label{equ:wg}
	^w\vect{g} = \, ^w\vect{H}_c \, ^c\vect{g}
	\end{equation}
	
\begin{figure}[t]
		
		\vspace{1em}
		\centering
		\includegraphics[width=0.51\linewidth]{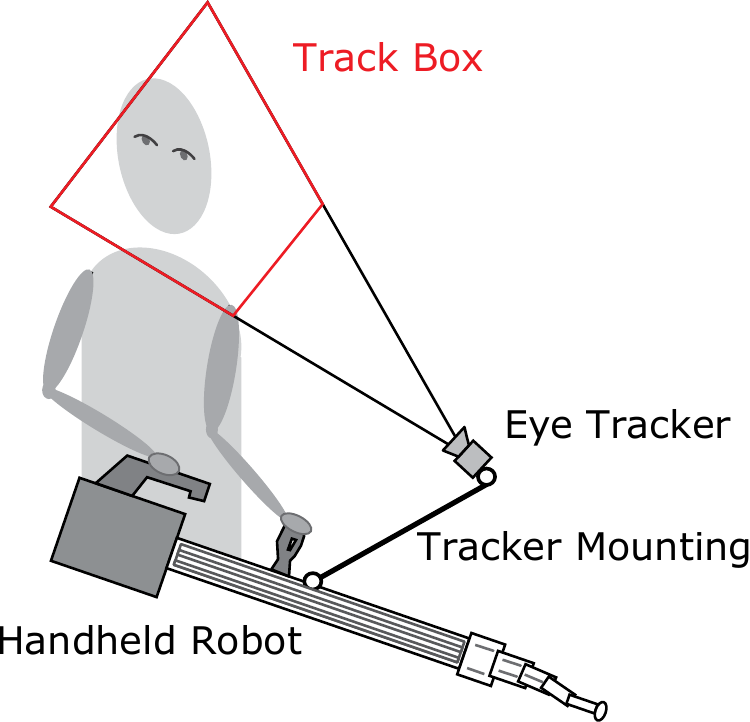}
		\caption{ Illustration of the handheld robot with the mounted eye tracker. The mount supports 2-DoF adjustment so that the user's head remains in the (red) trackable volume.}
		\label{fig:as-head-distance-to-track-box}
		\vspace{-0.68em}
	\end{figure}

	
	The eye tracker is mounted on to the handheld robot as it can be seen in figure \ref{fig:as-head-distance-to-track-box}. The position and orientation of the eye tracker can be adjusted so that the system can be adapted to varying user heights. In its generic configuration, the tracker is aligned such that it has the best accuracy within the robot's local workspace. Figure \ref{fig:front-profile-robot-labeled} shows a picture of the complete system. 
	
	A remote gaze tracker is preferred as it goes in line with the notion of a self-contained handheld tool. That way, users are not imposed to wear anything to use the tool which gets us closer to the handheld robot's aim of \textit{pick up and use}.
	
	\section{Eye Tracking Accuracy Study}
	The purpose of the accuracy study is to identify the limits of the introduced remote eye tracking system. Within the context of the handheld robot application, we are particularly interested in the constraints of the trackable area relative to the robot's workspace; that is how far the user can look away from the robot's tip for an accurate gaze capturing. Furthermore, we assess the accuracy of the measured gaze.
	
	\subsection{Data Collection}
	
	The general approach is to keep track of eye gaze data throughout a task where a participant would look and point (the robot tip) at a randomised sequence of targets which are broadly scattered over a workspace. In order to generate a broad variety of target sequences, three typical workspace setups were selected: The floor, a table surface and a vertical screen. 
	
\begin{figure}[h]
		\vspace{-1em}
		\centering
		\includegraphics[width=0.7\linewidth]{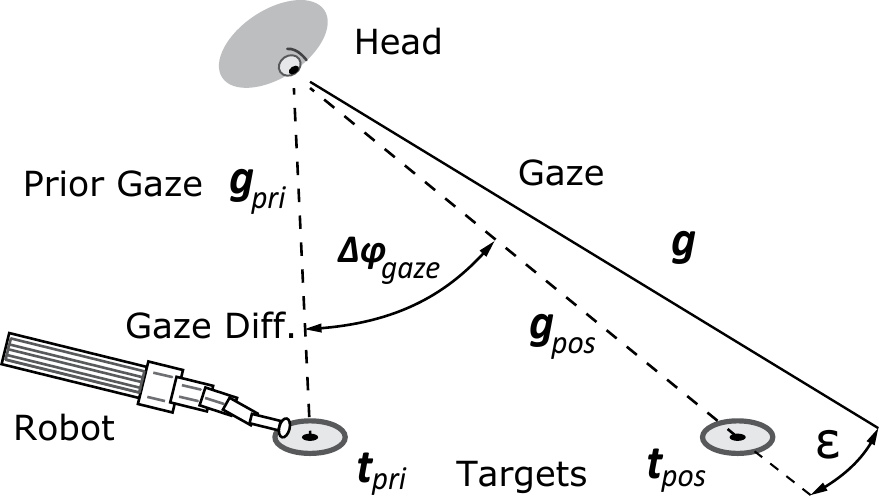}
		\caption{Illustration of a measurement iteration where a participant is proceeding from looking at a prior target $\vect{t}_{pri}$ to looking at the posterior target $\vect{t}_{pos}$ before moving the robot towards it. Dashed lines represent calculated true eye gaze rays whereas the solid line is the measured eye gazes.}
		\label{fig:asgazemeasurements}
		\vspace{-1em}
	\end{figure}

	For each target of the sequence, a participant is asked to look at a target without moving the robot and then touch the target with the robots end effector while looking at the target. This would then be the starting posture for the next target iteration. For each iteration, we keep track of the following data:
	
	{\small 
		\begin{itemize}[leftmargin=0.7in]
			\item[\textit{$\vect{g}_{pri}$ / $\vect{g}_{pos}$ }] the true eye gaze ray to a prior/posterior target
			\item[$\Delta\phi_{gaze}$] the angular gaze shift (difference) between posterior and prior true gaze
			\item[$\vect{g}$] the measured eye gaze
			\item[$\epsilon$] the angular error of the measured eye gaze
			\item[\texttt{Tracked}] True, if an eye gaze could be captured for a given target
		\end{itemize}
	}
	The relationship between those measurements can be seen in figure \ref{fig:asgazemeasurements}. $\vect{g}_{pri}$ and $\vect{g}_{pos}$ are obtained from the eye position which is known from a motion tracked helmet and the associated target of which the position is known too. The angular gaze shift $\Delta\phi_{gaze}$ is defined as \cite{Anonymous:PVm7qwCu}:

	\vspace{-1.5em}
	\begin{equation} \label{equ:phi}
	cos \Delta\phi_{gaze} = \frac{\vect{g}_{pri}\cdot\vect{g}_{pos}}{|\vect{g}_{pri}|\cdot|\vect{g}_{pos}|}
	\end{equation}
	
	For each targeting iteration, these measurements are taken for the case where the robot is pointed towards the prior target as well as when the robot tip touches the posterior target. 
	
	\subsection{Experiment Execution}
	%
	
	We recruited 11 participants for the 1st pilot gaze estimation experiment, mainly students from different fields (4 females, $M_{age} = 25$, $SD = 4.8$). Participation is on a voluntary basis as there is no financial compensation for their time. Each participant is asked to run through the aiming task for each target set-up of which the order is randomised. As part of their introduction, participants are given some practice time to familiarise themselves with the robot.
	
	For each of the three set-ups, we take the 2 data measurements for each target pair. Taking into account the measurement of the initial pose, we get 63 measurements per participant, so our final set contains 693 data points. 
	
	\subsection{Eye Gaze Modelling Results}
	
	In order to determine the accuracy performance of the eye tracking system, we split the data into the subsets $\vect{S}_{l}$ (N = 330), where the participant is looking at the next target and $\vect{S}_{p}$ (N = 363), where the target was aimed with the eye gaze and the robot's tip at the same time. These are further split to distinguish between the cases where the eye gaze was recognised (\texttt{Tracked = true}) or not which is denoted with an additional 1/0-index (1 = \texttt{true}). That way, we get the subsets $\vect{S}_{l,1}$, $\vect{S}_{l,0}$, $\vect{S}_{p,1}$ and $\vect{S}_{p,0}$ with sizes N = 174, 156, 331 and 32, respectively.
	
	$\vect{S}_{l}$ is used to investigate the effect of $\Delta\phi_{gaze}$ on $\epsilon$, while $\vect{S}_{p}$ is used to investigate the accuracy for the case where the user's gaze is close to the tip. For the analysis, data points with a difference to the mean higher than two standard deviations are removed so that 2.55\% is discarded.
	
\begin{figure}[t]
		
		\vspace{0.5em}
		\centering
		\includegraphics[width=0.99\linewidth]{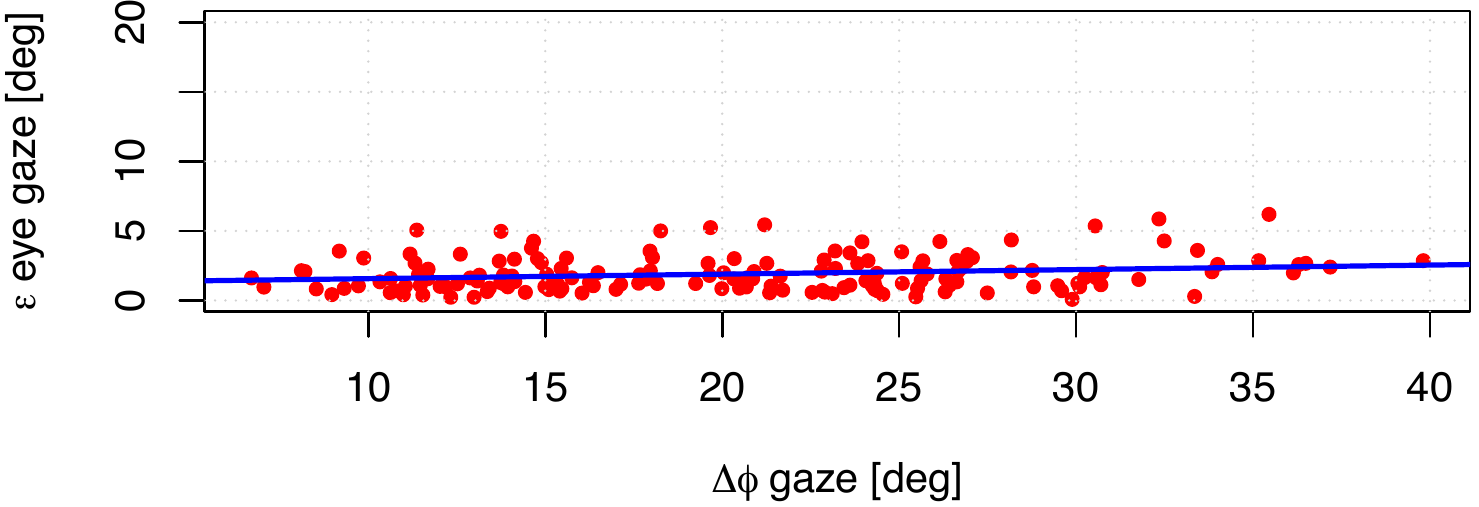}
		
		\vspace{-0.5em}
		\caption{Diagram of the linear regression model (blue) of the eye gaze error $\epsilon$ over the gaze shift $\Delta\phi_{gaze}$ (red samples).}
		\label{fig:aslinear-regression-eye-gaze-2}
		\vspace{-1em}
	\end{figure}

	For $\vect{S}_{p,1}$, we find a mean angular error of $\epsilon_0 = 1.99, CI[1.83, 2.16]$. The set $\vect{S}_{l,1}$ is analysed using a linear regression model.  We calculate the values $c_1 = 1.243$ and $c_2 = 0.032$ for the model of the shape
	\begin{equation}\label{equ:linear-model}
	\epsilon(\Delta\phi_{gaze}) = c_1 + c_2 \Delta\phi_{gaze}
	\end{equation}
	where the slope $c_2$ is significant ($p = .012, , R^2 = 0.032$). A diagram of the model can be seen in figure \ref{fig:aslinear-regression-eye-gaze-2}.
	
\begin{figure}[h]
		\vspace{-1em}
		\centering
		\includegraphics[width=0.99\linewidth]{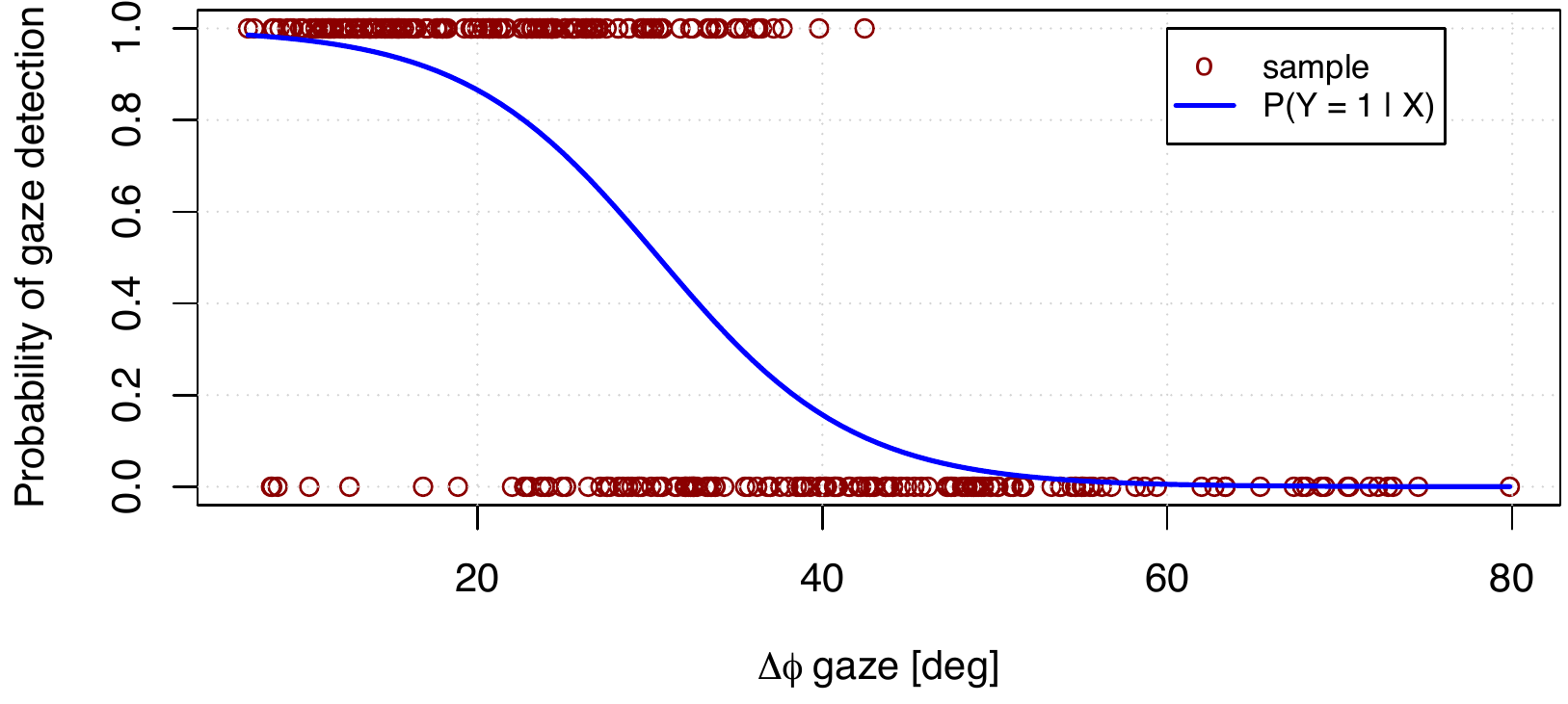}
		\caption{Diagram of the logit model (blue) to estimate the probability $P(Y = 1 | X)$ of successful eye tracking for a given gaze shift $\Delta\phi_{gaze}$. The (red) samples are the binary \texttt{Tracked} labels where \texttt{true} = 1 and \texttt{false} = 0.}
		\label{fig:aslogistic-regression-plot}
	\end{figure}

	In order to estimate the limit of the gaze shift angle $\Delta\phi_{gaze}$ for which gaze tracking delivers reliable results, we use $\vect{S}_{l}$, the whole subset of samples where the participant was not looking at the tip. A logistic regression \cite{Ziegler:2015jx} is performed using the model
	\vspace{-0.7em}
	\begin{equation}
	P(X) = \frac{\exp(\beta_{0} + \beta_{1}X) }{1 + \exp(\beta_{0} + \beta_{1}X)}
	\end{equation}
	where the independent variable is $\Delta\phi_{gaze}$ and $P(X)$ is the probability of the eye gaze being tracked (\texttt{Tracked = true}). 
	In order to choose the optimal threshold value as a decision point for the model, we run a 5-fold cross-validation over the range $P(X) \in [.25, .75]$.    
	As a result, we gain the decision point at $P(X) = 0.65$ for which the model fits 85.6\% of the data and we get the coefficients $\beta_0 = 5.407$ and $\beta_{1} = -0.177$ ($p < 2e-16 $ each) using the complete data set. By inverting the function at the decision point, we find that $P(X) > .65$ for $\Delta\phi_{gaze} \in [0, 27]$ (cf. figure \ref{fig:aslogistic-regression-plot}).\\
	
	Feeding the $\Delta\phi_{gaze}$ range back into the linear model (equation \ref{equ:linear-model}), we find $\epsilon_{27} = \epsilon(\Delta\phi_{gaze} = 27) = 2.107$ as an error prediction for the trackable range. 

	\subsection{Gaze tracking discussion}
	In addressing the question about the constraints of a workspace for eye tracking, it was found that it is limited by a maximum angle of \SI{27}{deg} away from the robot's end effector.
	Within this constraint, we anticipate a probability of successful eye tracking above 0.65 which informs subsequent studies in terms of the limitations of experimental designs. As the angular limit goes to any direction, the workspace has the shape of a cone with a tip angle of $2\times\SI{27}{deg} = \SI{54}{deg}$. 
	
	Concerning the angular accuracy of the eye tracking device in handheld robot applications, we identified a link to $\Delta\phi_{gaze}$. However, the linear regression yields a small slope coefficient indicating a small effect of the gaze direction on the error.
	The error for a range below \SI{27}{deg} is smaller than the maximum of the CI of the error of the $\vect{S}_{p}$ set. Therefore, an average error of up to $\epsilon = \SI{2.16}{deg}$ is anticipated.

	\section{The Attention Model}
	On the studies described in the rest of this paper, we apply our results from the eye tracking study for incorporating user attention estimation.
	
\begin{figure}[h]
		\centering
		\includegraphics[width=0.8\linewidth]{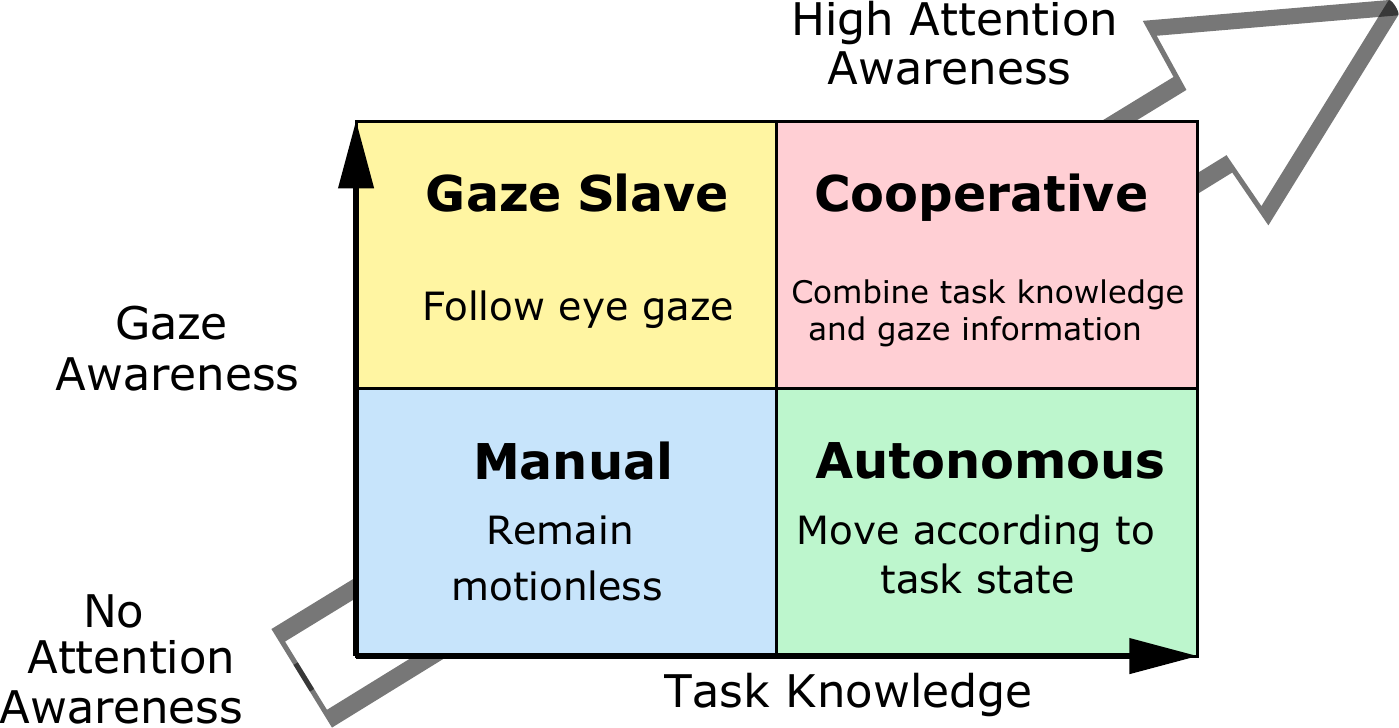}
		\caption{This behaviour diagram illustrates how the four behaviour modes of the robot are linked to the attention model which is based on the level of gaze awareness and task knowledge.}
		\label{fig:inv-behaviour-table}
	\end{figure}

	The attention model for the handheld robot is based on two factors: gaze awareness and task knowledge. These factors are inspired by the work by Land at al. who suggest that eye gaze is closely related to the location of a person's action \cite{Land:2016kw}. Therefore, we propose the following assumptions:
	\begin{enumerate}[label=A\arabic*]
		\item The part of a workspace which is watched by the user is within the user's focus of attention.  
		\label{A:area}
		\item A watched object is more likely to be in the user's focus of attention when it is task-relevant than when it is irrelevant to the task.
		\label{A:object}
	\end{enumerate}
	Based on these assumptions, we create a behaviour matrix with the gaze awareness and the task knowledge as the two axes which determine the behaviour. As the attention awareness is the product of both gaze awareness and task knowledge, it increases over the diagonal axis of the matrix as illustrated in figure \ref{fig:inv-behaviour-table}.

	The details about each behaviour mode are described in the following: 
	
	\begin{enumerate}[label=B\arabic*]
		\item \textbf{Manual Mode}: The robot remains motionless since neither the gaze nor the task knowledge influence the behaviour. 
		\label{B:Rigid}
		
		\item \textbf{Slave Mode}: In this mode, the robot ignores the status of an object or whether it is related to the task at all. Instead, the behaviour is purely determined by the estimation of the user's area of attention. This goes in line with assumption \ref{A:area} so that the robot follows the user's eye gaze in the workspace.
		\label{B:Slave}
		
		\item \textbf{Autonomous Mode}: The robot ignores any user actions and follows its own plan to complete the task. Choosing the sequence of task objects and finishing the job is fully automated.
		\label{B:Dominant}
		
		\item \textbf{Cooperative Mode}: The focus of attention is modelled as the intersection between the gazed at area and the location of a task-relevant object which goes in line with \ref{A:object}. The robot follows the eye gaze and helps to aim when a task object is focused. While the robot finishes the job, the user can shift the visual focus to a subsequent object. The robot catches up with the eye gaze once the task with the prior object is completed.
		\label{B:Cooperative}
	\end{enumerate}

	\section{Methodology of Attention Study}
	Having developed a set of behaviours which include our novel attention modelling, we assess the different modes through a gamified user study.
	
	The motivation for the game is driven by the demand for a task that would be easy enough for novice users to solve and which would be solvable using the handheld robot. At the same time, the robot would depend on being tactically reached by a user i.e. it could not solve the task by itself.

	\begin{figure*}[t!]
		\centering
		\begin{subfigure}[t]{0.32\textwidth}
			\centering
			\includegraphics[width=0.99\linewidth]{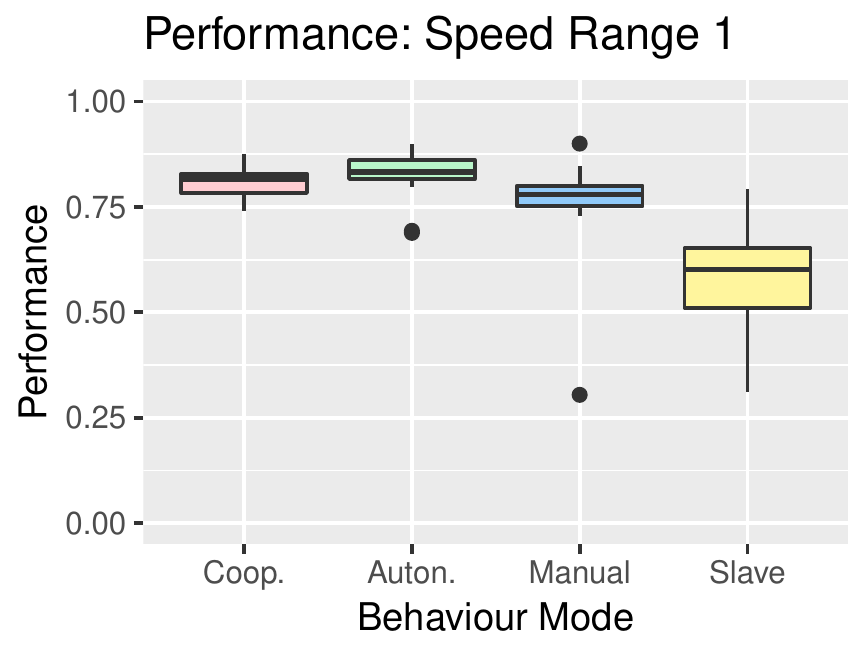}
		\end{subfigure}%
		~ 
		\begin{subfigure}[t]{0.32\textwidth}
			\centering
			\includegraphics[width=0.99\linewidth]{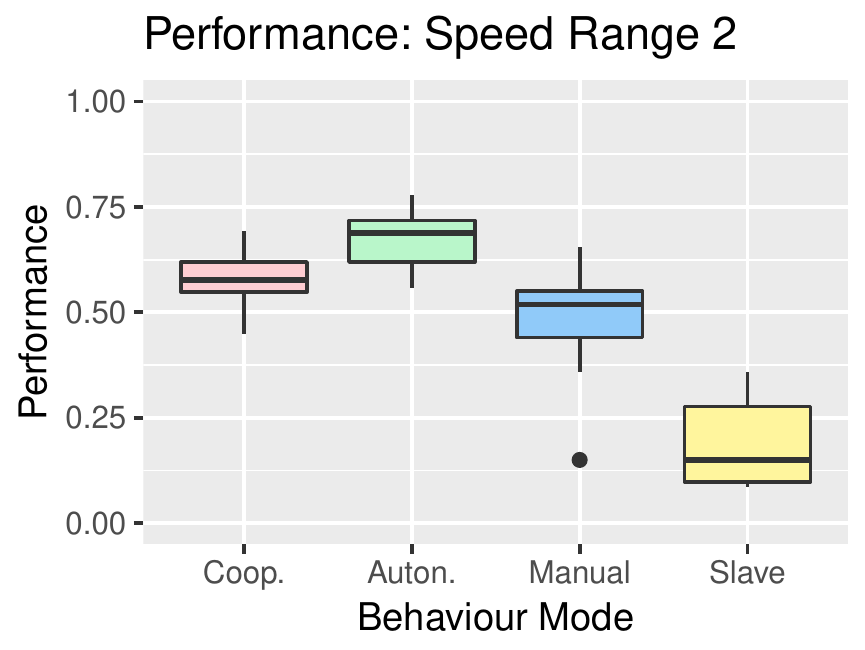}
		\end{subfigure}
		~ 
		\begin{subfigure}[t]{0.32\textwidth}
			\centering
			\includegraphics[width=0.99\linewidth]{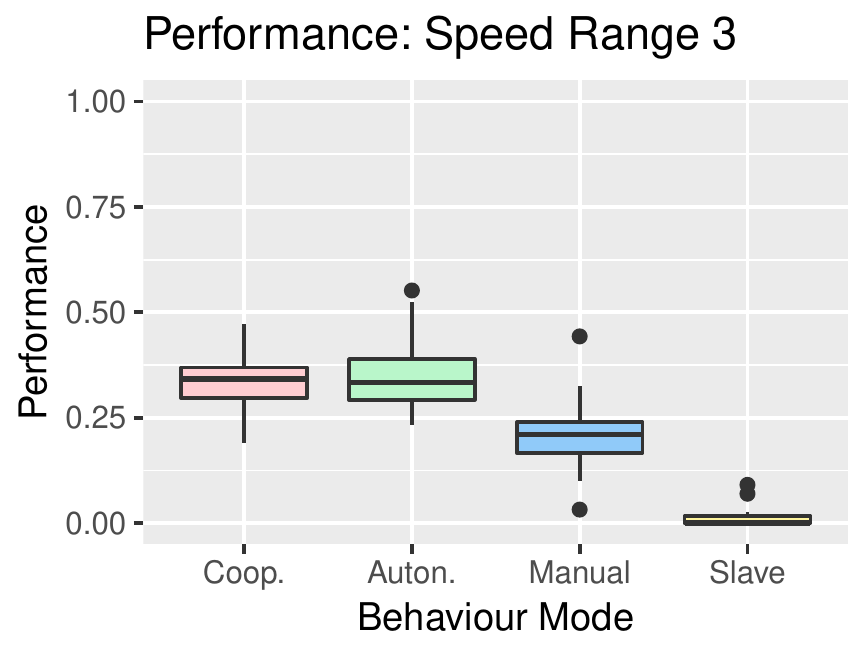}
		\end{subfigure}
		\caption{Performance (higher is better) measured in completed targets over total targets for each mode and speed range.}
		\label{fig:performanc}
	\end{figure*}

	\begin{table*}[t!]
		\centering
		\begin{subfigure}[t]{0.32\textwidth}
			\centering
			\includegraphics[width=0.9\linewidth]{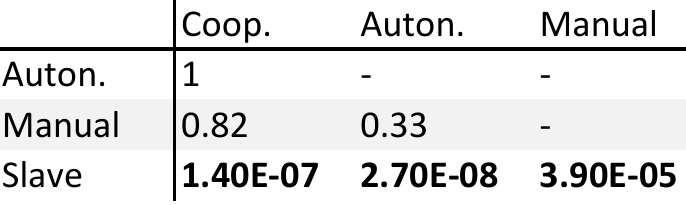}
			\caption{Speed Range 1}
			\label{tab:resultsttestr1}
		\end{subfigure}%
		~ 
		\begin{subfigure}[t]{0.32\textwidth}
			\centering
			\includegraphics[width=0.9\linewidth]{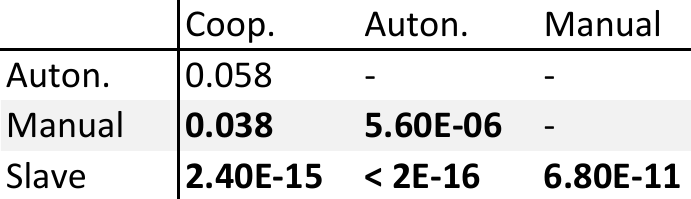}
			\caption{Speed Range 2}
			\label{tab:resultsttestr2}
		\end{subfigure}
		~ 
		\begin{subfigure}[t]{0.32\textwidth}
			\centering
			\includegraphics[width=0.9\linewidth]{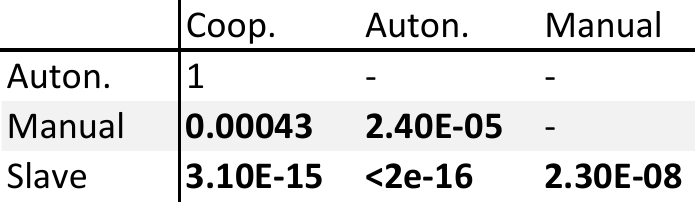}
			\caption{Speed Range 3}
			\label{tab:resultsttestr3}
		\end{subfigure}
		\vspace{1em}
		\caption{Bonferroni corrected $p$-values of pairwise t-test results for the 3 different speed ranges. Significant ($p < 0.05$) values are displayed in bold.}
		\label{tab:resultsttestr}
		\vspace{-1em}
	\end{table*}

	\subsection{Validation Task}    
	The game principle is inspired by \textit{Space Invaders}\footnote{Available for example at \textit{http://www.pacxon4u.com/space-invaders/}} and displayed on a 2D LCD screen (\SI{105}{\cm} diagonal). Targets travel with a constant speed from the upper edge of the screen to the bottom line. The aim of the game is to stop as many targets as possible before they reach the bottom. The robot's tip emits a virtual laser that can be used to stop a target. The laser needs to be activated via a trigger on the robot's handle and the tip needs to be close to the target ($< \SI{100}{\mm}$). It takes some lasering time to stop a target but less time for quicker targets, since otherwise, it would be impossible to complete it before it reaches the bottom line. The outcomes of the accuracy experiments are used to dimension both screen size and target diameter.
	
	The design of the game is grey scale coloured to avoid a disadvantage for colour blind people. The background of the game is a cartoon landscape and alongside the targets, there are similar objects dropping which cannot be stopped but are used as distractor stimuli. While 50\% of the targets are spawned randomly, the rest are part of a challenging scenario, for example, a triangle/line formation or an arrangement, where some slower targets are being taken over by quicker ones. This forces the player into situations where target priorities are equal or where plans have to be adapted to an unexpected event. An example of the gameplay can be seen in figure \ref{fig:invgame}.
	
	The robot, the screen and the eye tracker are tracked using an OptiTrack\footnote{http://optitrack.com} motion capturing system. Therefore, the intersection between the robot's tip and the screen, the user's eye gaze and the position of the targets is known in 3D space at all times. Note that the robot's task knowledge includes the state completion for each target; that way it is able to lock its orientation towards it and overwrite the laser trigger while it is being stopped in the cooperative and autonomous mode. However, while the robot decides the sequence in the autonomous mode, it is dependent on user attention in the cooperative mode (cf. B1-B4). 
	
	\subsection{Attention Experiment}
	
	For this new experiment for the attention study, we recruited 15 participants (6 females, $M_{age} = 25.5$, $SD = 5.6$). Many are students from technical courses, however, there is no expertise required to solve the task. There is no financial compensation for their time, however, many volunteers are thankful for trying the robot and the game and they are offered some refreshments. Each participant ran 3 game trials in each behaviour mode for a duration of \SI{80}{\s} each. They were exposed to targets with varying speeds and stopping times and the order of the behaviour modes was randomised to cancel out training effects. Before starting the experiment session, the participants were given an explanation and demonstration for each mode plus some practice time to get familiar with them.
	
	During trials, we keep track of the count of targets that are completed as well as the total number of targets that are presented. For the event of a target being stopped by the user or reaching the bottom line (e.g. the user failed to stop it), a data point is created which contains the target's speed and the current behaviour mode of the robot. Over the 160 trials, we collect 17k target samples in total.    
	
	The game runs with an update rate of \SI{60}{\Hz} and when one of the gaze-based modes is used (\ref{B:Slave} - \ref{B:Cooperative}), we register whether the tracker recognises the eye gaze for later analysis.
	
	After each trial, the participants are asked to fill out a questionnaire to assess the current trial and mode. The main parts of the questionnaire are the NASA TLX criteria \cite{Hart:1988hoa} which are used to measure the subject's task load. Furthermore, we asked participants to what extent they agree with the statements: \textit{The robot helped me with the task} and \textit{The robot obstructed me during the task} on a 5-point Likert scale (\textit{Strongly agree, Agree, Neither agree nor disagree, Disagree, Strongly disagree}). Also, they were given the chance to provide feedback comments for the current trial.
	
	\section{Results}
	The target data set is used to assess subject performance for each behaviour mode. Here, performance is defined as the proportion of completed targets over the total targets presented. Furthermore, the set is split into three speed ranges $R_{1,2,3}$: $[70, 200), [200, 330)$ and $[330, 490]$ (in mm/s) for a separate analysis. 2.1\% of the data points are outside of these ranges and are thus discarded. 
	
	\subsection{Mode Performance}
	
	The effect of the speed range and the behaviour mode on the performance is determined using a two-way factorial repeated measures ANOVA. As the results yield a significant effect for each factor ($p < 0.001$), they are further explored using post-hoc pairwise t-tests where Bonferroni correction is used. The mode-dependent differences in performance for each speed range can be seen in figure \ref{fig:performanc} and the associated t-test results are displayed in table \ref{tab:resultsttestr}.
	
	For every speed range, the slave mode is outperformed by the other modes and the performance yields a significant difference to each. The cooperative mode and the autonomous mode outperform the manual mode for each speed range, however, significance can only be determined for the two higher speed ranges $R_2$ and $R_3$ but not for $R_1$. In no case could a significant difference between the performance of the cooperative mode and the autonomous mode be found.
	
	We do not find any correlation between performance and age, gender, hours per week that video games are played or whether vision aids such as glasses or contact lenses were used.
	
	Considering the set of game update frames where the eye tracker did not recognise the eye gaze, we note that those add up to a share of 49.9\%. However, they are evenly distributed over the trial time so that in only 5.1\% of the time, these frames locally add up to over \SI{150}{ms}.

	\subsection{Task Load Index}
	
\begin{figure}[h]        
		\vspace{-1em}
		\centering
		\includegraphics[width=0.9\linewidth]{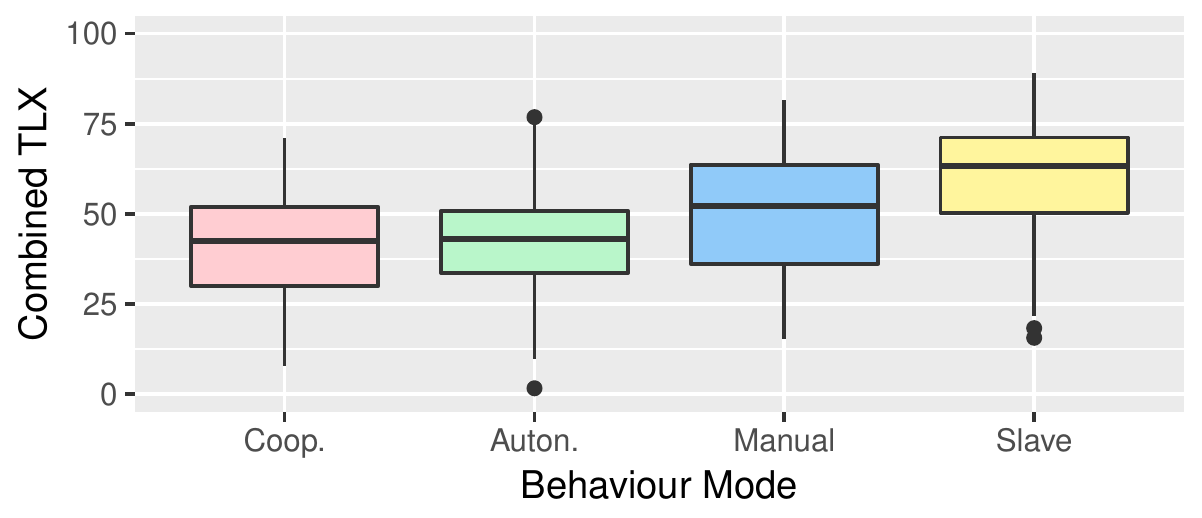}
		\caption{Perceived task load for each behaviour mode measured by the combined NASA TLX (lower is better).}
		\label{fig:resultscombinedtlxcombinedspeeds}
	\end{figure}

	For the analysis of the modes' effect on the perceived task load i.e. the combined NASA TLX results, we applied an ANOVA for the combined dataset and determine significance ($p< 0.001$). We proceed with a post-hoc pairwise t-test between the modes where the p-values (displayed in table \ref{tab:resultsttesttlx}) are Bonferroni corrected. The mode-dependent differences of TLX results can be seen in the diagram in figure \ref{fig:resultscombinedtlxcombinedspeeds}. The results for the slave mode yield a significant difference to the cooperative mode and to the autonomous mode only. Furthermore, the cooperative mode outperforms the manual mode with significance.
	
	\vspace{-1em}
	\begin{table}[h]
		\centering
		\includegraphics[width=0.6\linewidth]{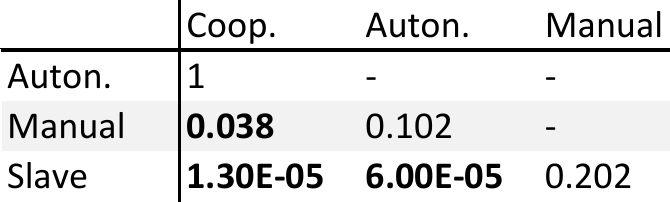}
		\caption{Bonferroni corrected $p$-values of pairwise t-test results for the mode depended mean differences of TLX outcomes. Significant ($p < 0.05$) values are displayed in bold.}
		\label{tab:resultsttesttlx}        
		\vspace{-1em}
	\end{table}
	
	\subsection{Helpfulness and Obstruction}
\begin{figure}[h]
		\vspace{0.5em}
		\centering
		\begin{subfigure}[t]{0.23\textwidth}
			\centering
			\includegraphics[width=0.99\linewidth]{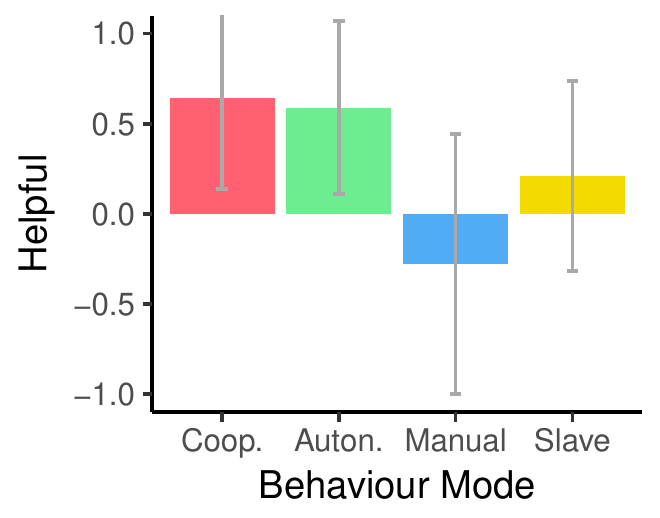}
			
			\vspace{-0.5em}
			\caption{Higher is better}
			\label{fig:resultshelpful}
		\end{subfigure}%
		~ 
		\begin{subfigure}[t]{0.23\textwidth}
			\centering
			\includegraphics[width=0.99\linewidth]{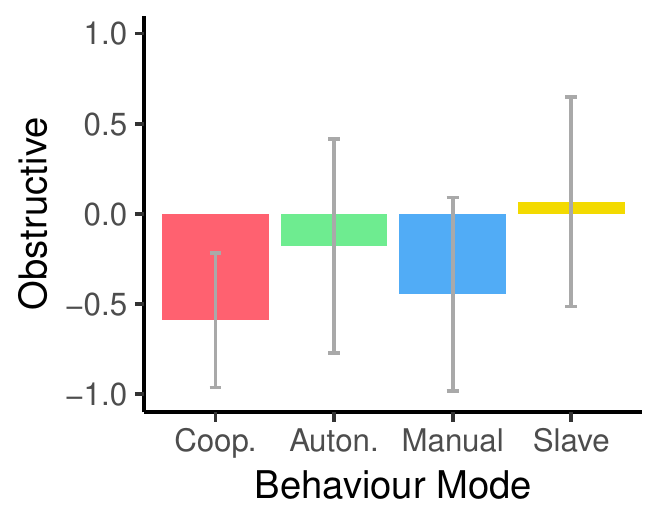}
			
			\vspace{-0.5em}
			\caption{Lower is better}
			\label{fig:resultsobst}
		\end{subfigure}
		\vspace{0.5em}
		\caption{Participant's rating of the robot being helpful (left) or obstructive (right) where 1 is \textit{strongly agree} and -1 is \textit{strongly disagree}. Whiskers indicate standard errors.}
		\label{fig:help.obst}
		\vspace{-1em}
	\end{figure}

	The 5-point Likert scales (from \textit{strongly disagree} to \textit{strongly agree}) for the statements about the robot's helpfulness and obstruction are scaled to numeric values on the interval $[-1, 1]$. As can be seen in figure \ref{fig:help.obst}a, the robot is rated most helpful in the cooperative (0.64) and autonomous (0,59) mode followed by the slave mode (0.21) while the manual mode tends towards unhelpful (-0.28). 
	\vspace{-1em}
	
	\subsection{Qualitative Feedback}
	As commenting on the trial behaviour was optional, the number varies for the different modes. We received many comments on the cooperative and autonomous mode and a few for the slave mode while the manual mode remained mostly uncommented on.
	
	The comments on the cooperative mode are mostly positive and often refer to collaboration experience e.g. \textit{I feel comfortable with the robot's assistance} or \textit{It feels like a team} and \textit{The robot helped me with accuracy once I chose a target}. Also, participants often pointed out that they felt in control e.g. \textit{I like the shared control} or \textit{I feel more in control with it}.
	
	For the autonomous mode, comments are positive when referring to accuracy e.g. \textit{The robot is better than me} and negative (mostly for fast targets) in terms of the robot's predictability e.g. \textit{I was irritated when the robot changed plans} and \textit{Why are you going there?}
	
	Within the slave mode, participants were complaining when eye tracking was faulty e.g. \textit{There was some offset, the robot did not follow accurately} and \textit{I saw a target but [the robot] did not follow}.
	
	The few comments on the manual mode were addressing physical workload (\textit{It was exhausting}).

	\section{Discussion}
	In addressing research question \ref{q1}, two gaze-based attention models where introduced where one additionally takes into account the robot's task knowledge. These modes were tested against the robot in a fully autonomous mode and for the scenario where the same job was done manually. 
	
	Regarding \ref{q2}, the modes were analysed with respect to performance and task load. 
	We found that using the primitive approach, where the robot is following the eye gaze, performance decreases while the workload is increased in comparison to the manual case. One explanation might be the lack of accuracy for eye tracking and that peripheral view could not be taken into account. Moreover, it was observed that the robot's motion towards the focus of gaze influenced the gaze behaviour which in turn caused more tip motion. This sometimes led to off-set errors and jittering during the use of the slave mode which was also reported in the feedback.
	
	In contrast, we found that the attention-based cooperative behaviour of the robot exceeds the manual analogue in terms of performance and decreases workload which makes it appear similar to the autonomous mode. This statement, however, is constrained to the requirement of a certain level of temporal demand for the effect to become apparent. The specific speed constraints for each mode are subject to further investigation. 
	
	When the cooperative mode is compared to fully autonomous, the statistics do not yield a difference in terms of task performance. However, we note that the autonomous mode is modelled with full omniscience which might not be possible outside the lab environment. For example, the robot might know which objects are task-relevant, while only the user knows the right sequence of task steps. Furthermore, qualitative feedback indicates that the robot is more predictable in the cooperative mode, making it more preferable.

	\section{Conclusion}
	This paper presents a system for estimating user attention for handheld collaboration. Gaze information and task knowledge are used as the two factors of the attention model. First, we developed a gaze model that estimates a 3D gaze ray from 2D gaze information and motion tracking. The gaze model was used to inform a subsequent attention study in which attention incorporated gaze based behaviours with varying levels of robot's task knowledge. The performance and task load in these modes were compared against a fully autonomous mode and a manual mode.
	
	Results indicate that cooperative behaviour is more effective than completing the task manually for the cases where there is a high demand for speed. We also found that task load is reduced when cooperative behaviour is based on both task knowledge and eye gaze i.e. with incorporated attention. 
	
	Moreover, the information of user attention is an essential step towards the prediction of intention. We suggest that our findings are used to inform future studies where the focus of attention can be used as part of a more sophisticated intention model. 
	We suggest that a handheld robot, enhanced with such a model, could be used in applications supporting users with varied skill in e.g. manufacturing as in the fields of assembly or welding where the attention model could help the robot choosing subtasks with respect to user preferences.
	
	{\bf Acknowledgement} This work was partially supported by the German Academic Scholarship Foundation and by the UK's Engineering and Physical Sciences Research Council. Opinions are the ones of the authors and not of the funding organisations.
	
	\vspace{-0.8em}

	
	\bibliographystyle{unsrt} 
	\bibliography{references}

\begin{thebibliography}{10}

\bibitem{GreggSmith:2015bh}
Austin Gregg-Smith and Walterio~W Mayol-Cuevas.
\newblock {The design and evaluation of a cooperative handheld robot}.
\newblock In {\em 2015 IEEE International Conference on Robotics and Automation
  (ICRA)}, pages 1968--1975. IEEE, 2015.

\bibitem{GreggSmith:2016hn}
Austin Gregg-Smith and Walterio~W Mayol-Cuevas.
\newblock {Investigating spatial guidance for a cooperative handheld robot}.
\newblock In {\em 2016 IEEE International Conference on Robotics and Automation
  (ICRA)}, pages 3367--3374. IEEE, 2016.

\bibitem{Land:2016kw}
Michael Land, Neil Mennie, and Jennifer Rusted.
\newblock {The Roles of Vision and Eye Movements in the Control of Activities
  of Daily Living}.
\newblock {\em Perception}, 28(11):1311--1328, 1999.

\bibitem{GreggSmith:2016cz}
Austin Gregg-Smith and Walterio~W Mayol-Cuevas.
\newblock {Inverse Kinematics and Design of a Novel 6-DoF Handheld Robot Arm}.
\newblock In {\em 2016 IEEE International Conference on Robotics and Automation
  (ICRA)}, pages 2102--2109. IEEE, 2016.

\bibitem{Vatsal:2017dy}
Vighnesh Vatsal and Guy Hoffman.
\newblock {Wearing Your Arm on Your Sleeve: Studying Usage Contexts for a
  Wearable Robotic Forearm}.
\newblock In {\em 2017 26th IEEE International Symposium on Robot and Human
  Interactive Communication (RO-MAN)}, pages 974--980. IEEE, 2017.

\bibitem{Echtler:2003uo}
Florian Echtler, Fabian Sturm, Kay Kindermann, Gudrun Klinker, Joachim Stilla,
  Jorn Trilk, and Hesam Najafi.
\newblock {The Intelligent Welding Gun: Augmented Reality for Experimental
  Vehicle Construction}.
\newblock In {\em Virtual and Augmented Reality Applications in Manufacturing},
  pages 333--360. Springer London, 2004.

\bibitem{Rivers:2012ff}
Alec Rivers, Ilan~E Moyer, and Fr{\'e}do Durand.
\newblock {Position-Correcting Tools for 2D Digital Fabrication}.
\newblock {\em ACM Transactions on Graphics (TOG)}, 31(4):88--7, August 2012.

\bibitem{Mansouryar:2016uv}
Mohsen Mansouryar, Julian Steil, Yusuke Sugano, and Andreas Bulling.
\newblock {3D Gaze Estimation from 2D Pupil Positions on Monocular Head-Mounted
  Eye Trackers }.
\newblock In {\em Proceedings of the Ninth Biennial ACM Symposium on Eye
  Tracking Research Applications}, pages 197--200. ACM, March 2016.

\bibitem{Leelasawassuk:2015cq}
Teesid Leelasawassuk, Dima Damen, and Walterio~W Mayol-Cuevas.
\newblock {Estimating Visual Attention from a Head Mounted IMU}.
\newblock In {\em Proceedings of the ACM International Symposium on Wearable
  Computers}, pages 147--150. ACM, September 2015.

\bibitem{Stiefelhagen:2004gr}
R~Stiefelhagen, C~Fogen, P~Gieselmann, H~Holzapfel, K~Nickel, and A~Waibel.
\newblock {Natural Human-Robot Interaction using Speech, Head Pose and Gestures
  }.
\newblock In {\em 2004 IEEE/RSJ International Conference on Intelligent Robots
  and Systems (IROS) (IEEE Cat. No.04CH37566)}, pages 2422--2427. IEEE, 2004.

\bibitem{Odobez:2007hk}
Jean-Marc Odobez and Sileye Ba.
\newblock {A Cognitive and Unsupervised Map Adaptation Approach to the
  Recognition of the Focus of Attention from Head Pose}.
\newblock In {\em 2007 IEEE International Conference on Multimedia and Expo},
  pages 1379--1382. IEEE, 2007.

\bibitem{Mulvey:2012di}
Fiona Mulvey and Michael Heubner.
\newblock {Eye Movements and Attention}.
\newblock In {\em Gaze Interaction and Applications of Eye Tracking}, pages
  129--152. IGI Global, 2012.

\bibitem{Land:2001hl}
Michael~F Land and Mary Hayhoe.
\newblock {In what ways do eye movements contribute to everyday activities?}
\newblock {\em Vision Research}, 41(25-26):3559--3565, November 2001.

\bibitem{Anonymous:2001fx}
Jeff~B Pelz and Roxanne Canosa.
\newblock {Oculomotor behavior and perceptual strategies in complex tasks}.
\newblock {\em Vision Research}, 41(25-26):3587--3596, 2001.

\bibitem{Mennie:2006fo}
Neil Mennie, Mary Hayhoe, and Brian Sullivan.
\newblock {Look-ahead fixations: anticipatory eye movements in natural tasks}.
\newblock {\em Experimental Brain Research}, 179(3):427--442, December 2006.

\bibitem{Briot:2015jo}
S{\'e}bastien Briot and Wisama Khalil.
\newblock {Homogeneous Transformation Matrix}.
\newblock In {\em Dynamics of Parallel Robots}, pages 19--32. Springer
  International Publishing, Cham, June 2015.

\bibitem{Anonymous:PVm7qwCu}
I~N Bronshtein, K~A Semendyayev, Gerhard Musiol, and Heiner M{\"u}hlig.
\newblock {\em {Handbook of Mathematics}}.
\newblock Springer Science {\&} Business Media, 2013.

\bibitem{Ziegler:2015jx}
Andreas Ziegler.
\newblock {An Introduction to Statistical Learning with Applications. R. G.
  James, D. Witten, T. Hastie, and R. Tibshirani (2013). Berlin: Springer. 440
  pages, ISBN: 978-1-4614-7138-7.}
\newblock {\em Biometrical Journal}, 58(3):715--716, December 2015.

\bibitem{Hart:1988hoa}
Sandra~G Hart and Lowell~E Staveland.
\newblock {Development of NASA-TLX (Task Load Index): Results of Empirical and
  Theoretical Research}.
\newblock {\em Advances in psychology}, 52:139--183, January 1988.

\end{thebibliography}
	
	
	

\end{document}